\documentclass{article}

\usepackage{arxiv}

\def\BibTeX{{\rm B\kern-.05em{\sc i\kern-.025em b}\kern-.08emT\kern-.1667em\lower.7ex\hbox{E}\kern-.125emX}}
    
\usepackage[utf8]{inputenc} 
\usepackage[T1]{fontenc}    
\usepackage{hyperref}       
\usepackage{url}            
\usepackage{booktabs}       
\usepackage{amsfonts}       
\usepackage{nicefrac}       
\usepackage{microtype}
\usepackage{algorithm}

\usepackage{booktabs}

\usepackage{epsfig,verbatim}
\usepackage{amsmath}
\usepackage{bm}
\usepackage{graphicx,psfrag,epsf}
\usepackage{enumerate}
\usepackage{amssymb}
\usepackage{lscape}
\usepackage{color}
\usepackage{url}
\usepackage{mathrsfs}
\def\singlespace{\def\baselinestretch{1}\@normalsize}

\newtheorem{lemma}{Lemma}
\newtheorem{theorem}{Theorem}

\numberwithin{equation}{section}

\renewcommand{\hat}{\widehat}
\def\singlespace{\def\baselinestretch{1}\@normalsize}

\makeatother

\DeclareMathOperator*{\argminB}{argmin}
\newcommand{\by}{\bm{y}}
\newcommand{\bee}{\bm{e}}

\newcommand{\bX}{\bm{X}}

\newcommand{\bx}{\bm{x}}

\newcommand{\bbeta}{\mbox{\boldmath$\beta$}}
\newcommand{\beps}{\mbox{\boldmath$\epsilon$}}

\makeatletter
\DeclareRobustCommand*\cal{\@fontswitch\relax\mathcal}
\makeatother

\def\newpage{\vfill\eject}

\def\today{\ifcase\month\or
	January\or February\or March\or April\or May\or June\or
	July\or August\or September\or October\or November\or December\fi
	\space\number\day, \number\year}

\newdimen\biblioindent    \biblioindent=30pt

 at 7truept

\def\bSigma{\bm{\Sigma}}

\def\sign{\mbox{sign}}

\newcommand{\be}{\begin{equation}}
\newcommand{\ee}{\end{equation}}
\newcommand{\beq}{\begin{equation}}
\newcommand{\eeq}{\end{equation}}
\newcommand{\beqn}{\begin{eqnarray}}
\newcommand{\eeqn}{\end{eqnarray}}
\newcommand{\beqnn}{\begin{eqnarray*}}
	\newcommand{\eeqnn}{\end{eqnarray*}}

\title{ET-Lasso: A New Efficient Tuning of Lasso-type Regularization for High-Dimensional Data}

\author{
  Songshan Yang \\
  The Pennsylvania State University\\
  University Park, PA \\
  \texttt{songshan.yang@gmail.com} \\
   \And
 Jiawei Wen \\
  The Pennsylvania State University\\
  University Park, PA \\
  \texttt{jkw5392@psu.edu} \\
   \And
 Xiang Zhan \\
  The Pennsylvania State University\\
  Hershey, PA\\
  \texttt{xyz5074@psu.edu} \\
     \And
 Daniel Kifer \\
  The Pennsylvania State University\\
  University Park, PA \\
  \texttt{dkifer@cse.psu.edu} \\
}

\begin{document}
\maketitle


%
\begin{abstract}
The $L_{1}$ regularization (Lasso) has proven to be a versatile tool to select relevant features and estimate the model coefficients simultaneously and has been widely used in many research areas such as genomes studies, 
finance, and biomedical imaging. 
Despite its popularity, it is very challenging to guarantee the feature selection consistency of Lasso especially when the dimension of the data is huge. One way to improve the feature selection consistency is to select an ideal tuning parameter. Traditional tuning criteria mainly focus on minimizing the estimated prediction error or maximizing the posterior model probability, such as cross-validation and BIC, which may either be time-consuming or fail to control the false discovery rate (FDR) when the number of features is extremely large. The other way is to introduce pseudo-features to learn the importance of the original ones. Recently, the Knockoff filter is proposed to control the FDR when performing feature selection. However, its performance is sensitive to the choice of the expected FDR threshold. Motivated by these ideas, we propose a new method using pseudo-features to obtain an ideal tuning parameter. In particular, we present the \textbf{E}fficient \textbf{T}uning of \textbf{Lasso} (\textbf{ET-Lasso}) to separate active and inactive features by adding permuted features as pseudo-features in linear models. The pseudo-features are constructed to be inactive by nature, which can be used to obtain a cutoff to select the tuning parameter that separates active and inactive features. Experimental studies on both simulations and real-world data applications are provided to show that ET-Lasso can effectively and efficiently select active features under a wide range of scenarios.  

\end{abstract}

%
%

\keywords{high-dimensional data \and Lasso \and automatic tuning parameter selection \and feature selection}

%
%
\maketitle

\section{INTRODUCTION}
\label{sec:intro}

High dimensional data analysis is fundamental in many research areas such as genome-wide association studies, finance, tumor classification and biomedical imaging  \cite{donoho2000high, fan2006statistical}. 
The principle of sparsity is frequently adopted and proved useful when analyzing high dimensional data, which assumes only a small proportion of the features contribute to the response (``active''). 
Following this general rule, penalized least square methods have been developed in recent years to select the active features and estimate their regression coefficients simultaneously. Among existing penalized least square methods, the least absolute shrinkage and selection operator (Lasso) \cite{tibshirani1996regression} is one of the most popular regularization methods that performs both variable selection and regularization, which enhance the prediction accuracy and interpretability of the statistical model it produces. Since then, many efforts have been devoted to develop algorithms in sparse learning of Lasso. Representative methods include but are not limited to \cite{beck2009fast, wainwright2009sharp, zhou2009thresholding, bach2008bolasso, reeves2013approximate, nesterov2013gradient, shalev2011stochastic, boyd2011distributed, friedman2007pathwise}.

Tuning parameter selection plays a pivotal role for identifying the true active features in Lasso. For example, it is shown that there exists an Irrepresentable Condition under which the Lasso selection is consistent when the tuning parameter converges to $0$ at a rate slower than $n^{-1/2}$ \cite{yu2014modified}. The convergence in $\ell_2$-norm is further established under a relaxed irrepresentable condition with an appropriate choice of the tuning parameter \cite{meinshausen2009lasso}. The tuning parameter can be computed theoretically but the calculation can be difficult in practice, especially for high-dimensional data.
In literature, cross-validation \cite{stone1974cross}, AIC \cite{akaike1974new} and BIC \cite{schwarz1978estimating} have been widely used for selecting tuning parameters for Lasso. The tuning parameters selected by a BIC-type criterion can identify the true model consistently under some regularity conditions, whereas AIC and cross-validation may not lead to a consistent selection \cite{wang2007tuning, wang2009shrinkage}.  

These criteria focus on minimizing the estimated prediction error or maximizing the posterior model probability, which can be computationally intensive for large-scale datasets. 
In high dimensional setting, cross-validation often results in models that are not stable in estimation. Estimation stability with cross-validation (ESCV) \cite{lim2016estimation} is an alternative approach to CV, with improved performance on estimation stability. ESCV constructs pseudo solutions by fitting sub-groups of dataset, and devises an estimation stability (ES) metric, which is a normalized sample variance of the estimates. ESCV selects the regularization parameter that is a local minimum of the ES metric, and leads to a smaller model than CV. Modified cross-validation criterion (MCC) is developed in \cite{yu2014modified} as an alternative criterion performed on each validation set. MCC aims to improve the model selection performance by reducing the bias induced by shrinkage. In general, their method selects a larger regularization parameter than Lasso. 

Recently, Barber and Candes \cite{barber2015controlling} proposed a novel feature selection method ``Knockoff'' that is able to control the false discovery rate when performing variable selection. This method operates by first constructing Knockoff variables (which are pseudo copies of the original variables) that mimic the correlation structure of the original variables, and then selecting features that are identified as much more important than their Knockoff copies, according to some measures of feature importance. However, Knockoff requires the number of features to be less than the sample size, which cannot be applied to high dimensional settings where the number of features is much larger than that of samples. In order to fix this, Model-X Knockoffs \cite{candes2018panning} is proposed to provide valid FDR control variable selection inference under the $p \ge n$ scenario. However, this method is sensitive to the choice of the expected FDR level, and it cannot generate a consistent solution for the model coefficients. Moreover, as will be seen from the simulation studies presented in Section 4.1, we notice that the construction complexity of the Knockoff matrix is sensitive to the covariance structure, and it is also very time consuming when $p$ is large. 
Motivated by both the literature of tuning parameter selection and pseudo variables-based feature selection, we propose the \textbf{E}fficient \textbf{T}uning of \textbf{Lasso} (\textbf{ET-Lasso}) which selects the ideal tuning parameter by using pseudo-features and accommodates high dimensional settings where $p$ is allowed to grow exponentially with $n$. The idea comes from the fact that active features tend to enter the model ahead of inactive ones on the solution path of Lasso. We investigate this fact theoretically under some regularity conditions, which results in selection consistency and a clear separation between active and inactive features. 
We further propose a cutoff level to separate the active and inactive features by adding permuted features as pseudo-features, which are constructed to be inactive and uncorrelated with signals; consequently they can help rule out tuning parameters that wrongly identify them as active. The idea of adding pseudo-features is inspired by \cite{luo2006tuning, wu2007controlling}, which proposed to add random features in forward selection problems. In our method, the permuted features are generated by making a copy of $\bX$ and then permuting its rows. In this way, the permuted features have the same marginal distribution as the original ones, and are not correlated with $\bX$ and $\by$. Unlike the Knockoff method, which selects features that are more important than their Knockoff copies, ET-Lasso tries to identify original features that are more important than all the permuted features. We show that the proposed method selects all the active features and simultaneously filters out all the inactive features with an overwhelming probability as $n$ goes to infinity and $p$ goes to infinity at an exponential rate of $n$. The experiments in Section 4 show that ET-Lasso outperforms other existing methods under different scenarios. As noted by one reviewer, the random permutation-based methods have also been studied in more general non-linear feature selection with random forest \cite{rudnicki2015all, nguyen2015unbiased, kursa2010feature, sandri2008bias}. The permuted features are introduced to correct the bias of variable selection. Some variable importance measures in node split selection are affected by both the variable importance and the variable characteristics. By comparing original features with their  permuted copies, the bias incurred by variable characteristics is eliminated.

%
%


The rest of this paper is organized as follows. In Section \ref{sec:motive}, we introduce the motivation and the model framework of ET-Lasso. In Section \ref{sec:theory}, we establish its theoretical properties. Then, we illustrate the high efficiency and potential usefulness of our new method both by simulation studies and applications to a number of real-world datasets in Section \ref{sec:experiment}. The paper concludes with a brief discussion in Section \ref{sec:conclusion}.

To facilitate the presentation of our work, we use $s$ to denote an arbitrary subset of $\{1, 2, \cdots, p\}$, which amounts to a submodel with covariates $\bX_s=\{X_j, j \in s\}$ and associated coefficients $\bbeta_s=\{\beta_j, j \in s\}$. $s_c$ is the complement of $s$. We use $\|\cdot\|_0$ to denote the number of nonzero components of a vector and $\vert s \vert$ to represent the cardinality of set $s$. We denote the true model by $s^*=\{j: \beta_j \neq 0\}$ with $\ \vert s^*\vert=\|\bbeta^*\|_0=k$.

\section{MOTIVATION AND MODEL FRAMEWORK}
\label{sec:motive}
\subsection{Motivation}

Consider the problem of estimating the coefficients vector $\bbeta$ from linear model
\beq \label{2.1}
\by=\bX \bbeta+ {\bm \epsilon},
\eeq
where $\by=(y_1, \cdots, y_n)^T$ is the response, $\bX=(\bx_1, \cdots, \bx_n)^T = (\bX_1, \cdots, \bX_p)$ is an $n\times p$ random design matrix with $n$ independent and identically distributed (IID) $p$-vectors $\bx_1, \cdots, \bx_n$. $(\bX_1, \cdots, \bX_p)$ correspond to $p$ features. $\bbeta=(\beta_1, \cdots, \beta_p)^T$ is the coefficients vector and $\beps=(\epsilon_1, \cdots, \epsilon_n)$ is an $n$-vector of IID random errors following sub-Gaussian distribution with $E(\epsilon_i)=0$ and $Var(\epsilon_i)=\sigma^2$. For high dimensional data where $p > n$, we often assume that only a handful of features contribute to the response, i.e, $\vert s^* \vert = k \ll p$. 


We consider the Lasso model that estimates $\bbeta$ under the sparsity assumption. The Lasso estimator is given by
\beq\label{2.3}
\hat{\bbeta}(\lambda)=\argminB_{\bbeta} \frac{1}{2n}\|\by-\bX\bbeta\|^2_2+\lambda\sum\limits_{j=1}^{p}|\beta_j|,
\eeq 
where $\lambda>0$ is a regularization parameter that controls the model sparsity. Consider the point $\lambda$ on the solution path of (\ref{2.3}) at which feature $\bX_j$ first enters the model,
\beq\label{def}
Z_j=\sup \{\lambda: \hat{\beta}_j(\lambda) \neq 0  \}, 
\eeq
which is likely to be large for most of active features and small for most inactive features. Note that $Z_j$ accounts for the joint effects among features and thus can be treated as a joint utility measure for ranking the importance of features. For orthonormal designs, the closed form solution of (\ref{2.3}) \cite{tibshirani1996regression} for Lasso directly shows that 
\beq 
\min\limits_{j \in s^*} Z_j > \max\limits_{j\in s^*_c} Z_j. 
\eeq

In section 3, under more general conditions, we will show that
\beq \label{property}
P(\min\limits_{j\in s^*} Z_{j} > \max\limits_{j \in s^*_c} Z_{j})\rightarrow 1, 	 \ \ \text{as} \ n \rightarrow \infty.
\eeq
Property \eqref{property} implies a clear separation between active and inactive features. The next step is to find a practical way to estimate $\max_{j\in s^*_c} Z_j$ in order to identify active features, i.e., obtain an ideal cutoff to separate the active and the inactive features.  

\subsection{Model Framework}

\label{sec:pfc}
Motivated by property \eqref{property}, we calculate the cutoff that separates the active and inactive features by adding pseudo-features. Since pseudo-features are known to be inactive, we can rule out tuning parameters that identify them as active. 
 
The permuted features matrix ${\bX}^{\pi}=(\bx_{\pi(1)}, \bx_{\pi(2)}, \cdots, \bx_{\pi(n)})^T$, where $\{\pi(1), \pi(2),  \cdots, \pi(n) \} $ is a permutation of $\{1, 2, \cdots, n\}$, are used as the pseudo-features. In particular, matrix $\bX^{\pi}$ satisfies
\beq
\bX^{\pi^T}\bX^{\pi} = \bX^{T}\bX.
\eeq
That is, the permuted features possess the same correlation structure as the original features, while breaking association with the $\by$ due to the permutation. Suppose that the features $\bX_j$ are centered, then the design matrix $[\bX, \bX^{\pi}]$ satisfies
\beq
E_{\pi}[\bX, \bX^{\pi}]^{T}[\bX, \bX^{\pi}]=
\begin{bmatrix}
    \bSigma & 0 \\
    0 & \bSigma
   \end{bmatrix}
\eeq
where $\bSigma$ is the correlation structure of $\bX$, and the approximately-zero off-diagonal blocks arise from the fact that $E_{\pi}[\bX_{i}^T\bX_j^{\pi}] = 0$ when the features are centered.

Now  we define the augmented design matrix $\bX^{A}=[\bX, \bX^{\pi}]=[X^{A}_1, X^{A}_2, \cdots, X^{A}_p, X^{A}_{p+1}, \cdots, X^{A}_{2p} ]$, where $[X^{A}_1, X^{A}_2, \cdots, X^{A}_p]$ is the original design matrix and $[X^{A}_{p+1}, \cdots, X^{A}_{2p}]$ is the permuted design matrix. The augmented linear model with $\bX^A$ as design matrix is 
\beq\label{2.9}
\by=\bX^A\bbeta^{A}+\epsilon^A,
\eeq
where $\bbeta^{A}$ is a $2p$-vector of coefficients and $\epsilon^A$ is the error term. 
The corresponding Lasso regression problem is
\beq\label{2.10}
\hat{\bbeta}^A(\lambda) = \argminB_{\bbeta^A} \frac{1}{2n}||\by-\bX^A\bbeta^A||_2^2+ \lambda\sum\limits_{j=1}^{2p} (|\beta_j^A|).
\eeq
Similar to $Z_j$, we define $Z_j^A$ by 
\beq
Z_j^A=\max \{ \sup\{\lambda: \hat{\beta}^A_j(\lambda) \neq 0\}, 0 \} ,  j=1, 2,  \cdots, 2p, 
\eeq 
which is the largest tuning parameter $\lambda$ at which $X^A_j$ enters the model (\ref{2.9}). Since $\{X^A_j: j=p+1, \cdots, 2p\}$ are truly inactive by construction, by Theorem \ref{Thm1} in Section 3, it holds in probability that $\min_{j \in s^*} Z_j^A > \max_{p+1\leq j \leq 2p}Z_j^A$. Define $C_p=\max_{p+1\leq j \leq 2p}Z_j^A$, then $C_p$ can be regarded as a benchmark to separate the active features from the inactive ones. This leads to a soft thresholding selection
\beq\label{soft}
\hat{s}_{\pi}^{s}=\{j: Z_j^A >C_p, 1\leq j \leq p\}.
\eeq

We implement a two-stage algorithm in order to reduce the false selection rate. We first generate permuted features $\bX^{\pi_i}=(\bx_{\pi_i (1)}, \bx_{\pi_i(2)}, \cdots, \bx_{\pi_i (n)})^T, i=1, 2$. In the first stage, we select the $\hat{s}_{\pi_1}$ based on the rule (\ref{soft}) using $\bX^{\pi_1}$. Then in the second stage, we combine $\bX_{\hat{s}_{\pi_1}}$ and $\bX^{\pi_2}$ to obtain $\bX^{A_2}=[\bX_{\hat{s}_{\pi_1}},\bX^{\pi_2}]$ and select the final feature set $\hat{s}$. The procedure of ET-Lasso is summarized in Algorithm \ref{TF}.
\begin{algorithm} 
	\caption{ET-Lasso}
	\label{TF}
	\begin{itemize}
		\item[\textbf{1}.] Generate two different permuted predictor samples $\bX^{\pi_i}=(\bx_{\pi_i (1)}, \bx_{\pi_i(2)}, \cdots, \bx_{\pi_i (n)})^T, i=1, 2$ and then combine $\bX^{\pi_1}$ with $\bX$ to obtain  augmented design matrix $\bX^{A_1}=[\bX,\bX^{\pi_1}]$.  
		\item[\textbf{2}.] For design matrix $\bX^{A_1}$, we solve the problem 
	\beq\label{2.11}
	\hat{\bbeta}^{A_1}(\lambda) = \argminB_{\bbeta^{A_1}} \frac{1}{2n}||\by-\bX^{A_1}\bbeta^{A_1}||_2^2+ \lambda  \sum\limits_{j=1}^{2p} (|\beta_j^{A_1}|)
	\eeq
	over the grid $\lambda_1 > \lambda_2> \cdots> \lambda_d$. $\lambda_1=\max_j |(\bX^{A_1}_{j})^T \by|$ is the smallest tuning parameter value at which none of the features could be selected. $\lambda_d$ is the cutoff point. In other words, $\lambda_d$ can be regarded as an estimator of $C_p$. Then we use selection rule (\ref{soft}) to obtain $\hat{s}_{\pi_1}$.  
		\item[\textbf{3}.] Combine $\bX^{\pi_2}$ with $\bX_{\hat{s}_{\pi_1}}$, which only includes features in $\hat{s}_{\pi_1}$, to obtain the augmented design matrix $\bX^{A_2}=[\bX_{\hat{s}_{\pi_1}},\bX^{\pi_2}]$. Repeat Step 2 for the new design matrix $\bX^{A_2}=[\bX_{\hat{s}_{\pi_1}},\bX^{\pi_2}]$ over $\lambda_1^{'} > \lambda_2^{'}> \cdots> \lambda_d^{'}$ to select $\hat{s}$.
	\end{itemize}
\end{algorithm}


\textbf{Remark}. Apparently, more iterations in ET-Lasso would have a better control on false discoveries. The reason for adopting a two step method is to keep a balance between recall and precision. The asymptotic results of Theorem \ref{Thm3} indicate the two step method can effectively control the false discovery rate. 

\subsection{Comparison with ``Knockoff''}
The Knockoff methods have been proposed  to control the false discovery rate when performing variable selection \cite{barber2015controlling,candes2018panning}. Specifically the Knockoff features $\Tilde{\bX}$ obey
\beq
\Tilde{\bX}^T\Tilde{\bX} = \bSigma, \bX^T\Tilde{\bX} = \bSigma - diag\{s\},
\eeq
where $\bSigma = \bX^T \bX$ and $s$ is a p-dimensional non-negative vector. That is, $\Tilde{\bX}$ possesses the same covariance structure as $\bX$. 
The authors then set 
\beq 
W_j = \max(Z_j, \tilde{Z}_j)\times\sign{(Z_j - \tilde{Z}_j)}
\eeq
as the importance metric for feature $j$ and a data-dependent threshold $T$
\beq 
T = \min\{ t \in \mathscr{W}:\frac{\#\{j: W_j \leq -t\}}{\max(\#\{j: W_j \geq t\},1)} \le q \},
\eeq
where $\mathscr{W} = \{|W_j|: j=1, \cdots, p\}$ and $q \in (0,1)$ is the expected FDR level.
The Knockoff selects the feature set as $\{j: W_j \ge T\}$, which has been shown to have FDR controlled at $q$ \cite{barber2015controlling,candes2018panning}.

The main difference between Knockoff and ET-Lasso is that Knockoff method selects features that are clearly better than their Knockoff copies, while ET-Lasso method selects the features that are more important than all the pseudo-features. Compared with Knockoff, our method of constructing the pseudo-features is much simpler than creating the Knockoff features. Particularly, when the dimension of the data is extremely large, it is very time consuming to construct the Knockoff copies for each feature. On the other hand, the Knockoff method is not able to provide a consistent estimator for the model coefficients. In addition, the feature selection performance of Knockoff is sensitive to the choice of expected FDR ($q$) as shown by our experiments, and our method does not include hyper-parameters that need to be tuned carefully. The two step method within ET-Lasso can fix the vulnerability of using permuted sample when relevant and irrelevant variables are correlated as discussed in Knockoff paper \cite{barber2015controlling}. This can be shown from the comprehensive numerical studies in Section \ref{sec:experiment}.

\section{THEORETICAL PROPERTIES}
\label{sec:theory}
Property \eqref{property} is the guiding principle of our selection procedure that applies ET-Lasso to select the ideal regularization parameter. 
Now we study (\ref{property}) in a more general setting than orthonormal designs. We introduce the regularity conditions needed in this study.
\begin{itemize}
	\item[(C1)] (\textbf{Mutual Incoherence Condition}) There exists some $\gamma \in (0, 1]$ such that 
	\[
	\| \bX_{s^*_c}^T \bX_{s^*} (\bX_{s^*}^T \bX_{s^*})^{-1} \|_{\infty}\leq (1-\gamma), 
	\]
	where $\| M \|_{\infty}=\max\limits_{i=1, \cdots, m}\sum^{n}_{j=1}|M_{ij}|$ for any matrix $M=(M_{ij})_{m\times n}$.
	\item[(C2)] 
	There exists some $c_{\min} >0$ such that
	\[ 
	\lambda_{\min}(n^{-1}\bX_{s^*}^T\bX_{s^*})>c_{\min},
	\]
	where $\lambda_{\min}(A)$ denotes the minimum eigenvalue of A.
	\item[(C3)] $\log p=n^{\delta_1}$, $\vert s^* \vert=k= O(n^{\delta_2})$ and $\min\limits_{j \in s^*}|\beta_j|^2> O(n^{\delta_3-1})$, where $\delta_1, \delta_2>0$, and $\delta_1 + \delta_2 < \delta_3 <1$. 
\end{itemize}  
Condition (C1) is called mutual incoherence condition, and it has been considered in the previous work on Lasso \cite{wainwright2009sharp, fuchs2005recovery, tropp2006just}, that guarantees that the total amount of an irrelevant covariate represented by the covariates in the true model is
not to reach 1.
Condition (C2) indicates that the design matrix consisting of active features is full rank. Condition (C3) states some requirements for establishing the selection consistency of the proposed method. The first one assumes that $p$ diverges with $n$ up to an exponential rate, which allows the dimension of the data to be substantially larger than the sample size. The second one implies that the number of active features $k$ is allowed to grow with sample size $n$ but $k/n \rightarrow 0$ as $n \rightarrow \infty$. We also require the minimal component of $\bbeta_{s^*}$ does not degenerate too fast. 

One of the main results of this paper is that under (C1) - (C3), property \eqref{property} holds in probability:
\begin{theorem}\label{Thm1}
 Under conditions C1 - C3, assume that the design matrix $\bX$ has its $n$-dimensional columns normalized such that $n^{-1/2}\max_{j \in \{1,2,\cdots, p\}}\|X_j\|_2\leq 1$, then
	$$
	P(\min\limits_{j\in s^*} Z_{j} > \max\limits_{j \in s^*_c} Z_{j})\rightarrow 1, 	 \ \ \text{as} \ n \rightarrow \infty.
	$$

\end{theorem}
Theorem \ref{Thm1} justifies using $Z_j$ to rank the importance of features. In other words, $Z_j$ ranks an active feature above an inactive one with high probability, and thus leads to a separation between the active and inactive features. The proof is given in the supplementary material. 

The following result gives an upper bound on the probability of recruiting any inactive feature by ET-Lasso, and implies that our method excludes all the inactive features asymptotically when $n \rightarrow \infty $.  
\begin{theorem}\label{Thm3}
	Let $r$ be a positive integer, 
	by implementing the ET-Lasso procedure, we have  
	\beq \label{thm3.2}
	P(|\hat{s} \cap s^*_c| \geq 1)\leq O(exp(-n^{\delta_2})),
	\eeq
where $k= O(n^{\delta_2})$ as specified in condition (C2).  

\begin{proof} 
$\forall  \hat{s}_{\pi_i}$ and a fixed number $r$, the probability that 
\[
P(|\hat{s}_{\pi_i} \cap s^*_c|\geq r)=\frac{(p-k)!}{(p-k-r)!}(2p-k-r)!/(2p-k)!\leq (1-\frac{r}{2p-k})^p.
\]
Thus, 
\[
P(|\hat{s} \cap s^*_c| \geq r) \leq (1  - \frac{r}{2p-k})^{2p}.
\]
As $p \rightarrow \infty$, $k=o(p)$ and set $r=1$, 
\[
[(1-\frac{1}{2p-k})^{2p-k}]^k \rightarrow \exp(-k),
\] 
then we can claim that the upper bound for $P(|\hat{s} \cap s^*_c| \geq 1)$ is $\exp(-n^{\delta_2})$.
\end{proof}

\end{theorem}


Theorem \ref{Thm3} indicates that the number of false positives can be controlled better if there are more active features in the model, and our simulation results in Section 4 support this property.




\section{EXPERIMENTS}
\label{sec:experiment}
\subsection{Simulation Study}

In this section, we compare the finite sample performance of ET-Lasso with Lasso+BIC (BIC), Lasso+Cross-validation (CV), Lasso+ESCV and Knockoff (KF) under different settings. For CV method, 5-folded cross validation is used to select the tuning parameter $\lambda$. 
We consider three FDR thresholds for Knockoff, 0.05, 0.1 and 0.2, so as to figure out the sensitivity of its performance to the choice of the FDR threshold. The response $y_i$ is generated from the linear regression model (\ref{2.1}), where $\bx_i \sim N(0, \bSigma)$, $\epsilon_i \sim N(0, 1)$ for $i=1, 2, \cdots, n $. 
\begin{itemize}
\item[1.] the sample size $n=500$; 
\item[2.] the number of predictors $p=1000, 2000$;
\item[3.] the following three covariance structures of $\bm{X}$ considered in \cite{fan2008sure} are included to examine the effect of covariance structure on the performance of the methods:
\begin{description}
\item{(i)} Independent, i.e, $\bSigma = \bm{I}$,	
\item{(ii)} AR(1) correlation structure: $\bSigma =(\sigma_{ij})_{p \times p}, \sigma_{ij}=0.5^{|i-j|}$,
\item{(iii)} Compound symmetric correlation structure (CS): $\bSigma =(\sigma_{ij})_{p \times p}, \sigma_{ij}=1$ if $i=j$ and $0.25$ otherwise;
\end{description}
\item[4.] $|s^*|\equiv k=10, 15$, $\beta_{j}=(-1)^u\times 2$ for $j \in s^*$, where $u\sim$ Bernoulli (0.5), and $\beta_{j}=0$ for $j \in s^*_c$.

\end{itemize}
The simulation results are based on $1000$ replications and the following criteria are used to evaluate the performance of ET-Lasso:
\begin{itemize}
	\item[1.] ${\cal P}$: the average precision (number of active features selected/ number of features selected);
	\item[2.] $\cal R$: the average recall (number of active features selected/number of active features);
	\item[3.] ${\cal F}_1$: the average $F_1$-score (harmonic mean of precision and recall);
	\item[4.] Time: the average running time of each method.
\end{itemize}

The simulation results are summarized in Table \ref{tab:metrics} and \ref{tab:cs}. We can observe that ET-Lasso has higher precision and $F_1$ score than other methods under all circumstances. For independent setting, all methods except KF(0.05) successfully recover all active features, as suggested by the recall values. The average precision values of ET-Lasso are all above $0.97$, while Lasso+BIC has precision values around $0.6$, and Lasso+CV has precision values around $0.2$. Lasso+ESCV has precision values lower than $0.95$ for $k=10$, and lower than $0.88$ for $k=15$. KF(0.05) barely selects any feature into the model due to its restrictive FDR control, resulting in very small values in recall, and the numbers of selected features are zero in some of the replications. KF(0.1) and KF(0.2) successfully identify all active features, whereas their precision values and $F_1$ scores are smaller than ET-Lasso. The results for AR(1) covariance structure are similar to those of independent setting. In CS setting, KF based methods sometimes select zero feature into the model, and thus the corresponding precision and $F_1$ scores cannot be computed. ET-Lasso again outperforms other methods. In addition, ET-Lasso enjoys favorable computational efficiency compared with Lasso+CV, Lasso+ESCV and Knockoff. ET-Lasso finishes in less than $0.5$s in all settings, while Knockoffs require significantly more computing time, and their computational costs increase rapidly as $p$ increases. In addition, the performances of Knockoff rely on the choice of the expected FDR. When the correlations between features are strong, Knockoff method needs higher FDR thresholds to select all the active variables. 
\begin{table*}[htbp]
\vskip 0.15in
\centering
    \caption{
    Simulation results of ET-Lasso, Lasso+BIC, Lasso+CV and Knockoff with different FDR thresholds in independent and AR(1) covariance structure settings. Numbers in parentheses denote the corresponding standard deviations over the 1000 replicates. \# indicates an invalid average precision or $F_1$ score for methods that select zero feature in some of the replications.}
    \label{tab:metrics}
    \scalebox{1}{
      \begin{tabular}{lcccccccc}
        \hline
 &\multicolumn{4}{c}{Independent}  &\multicolumn{4}{c}{AR(1)} \\
 \hline
 & ${\cal P}$ & ${\cal R}$ & ${\cal F}_1$& Time  & ${\cal P}$ & ${\cal R}$ & ${\cal F}_1$& Time  \\ 
    \hline

     &\multicolumn{8}{c}{$p=1000$, $k=10$} \\

 \hline
ET-Lasso &0.97  & 1.0  & \textbf{0.98 (0.001)} & 0.27 (0.002)
& 0.93  & 1.0 & \textbf{0.96 (0.001)}& 0.27 (0.002) 
 \\

BIC &0.68  & 1.0  & 0.80 (0.003) & 0.09 (0.001) 
& 0.64  &1.0 & 0.77 (0.003) & 0.10 (0.001)
\\ 

CV & 0.20  & 1.0  & 0.33 (0.003) & 1.01 (0.006)
& 0.20  & 1.0  & 0.32 (0.003) & 1.01 (0.006)
\\

ESCV & 0.94  & 1.0  & 0.96 (0.002) & 2.33 (0.012)
& 0.92  & 1.0  & 0.95 (0.001) & 2.31 (0.012)
\\

KF(0.05) & \# & 0.00  &\# & 348.6 (12.10) 
&  \# & 1.0 &  \# & 427.8 (12.30)
 \\ 

KF(0.1)& 0.92  & 1.0  &0.96 (0.002) & 356.1 (13.01) 
& 0.91   & 1.0  & 0.95 (0.002) & 436.9 (13.21) 
\\ 

KF(0.2)&  0.83  & 1.0  & 0.90 (0.003) & 352.5 (12.23) 
& 0.82  & 1.0  & 0.89 (0.003) & 432.3 (13.07) 
\\
   \hline

    &\multicolumn{8}{c}{$p=1000$, $k=15$} \\
            \hline

ET-Lasso& 0.97  & 1.0  & \textbf{{0.99 (0.001)}} & 0.26 (0.001) 
&  0.94  & 1.0   & \textbf{0.97 (0.001)} & 0.27 (0.002)
\\ 

BIC & 0.63  & 1.0  & 0.77 (0.003) & 0.09 (0.001)  
& 0.59  & 1.0   &0.74 (0.003) & 0.09 (0.001) 
\\ 

CV & 0.21  & 1.0  & 0.34 (0.003) & 0.93 (0.004) 
&  0.20  & 1.0   &  0.33 (0.002) & 1.02 (0.006)  
\\ 

ESCV & 0.88  & 1.0  & 0.93 (0.002) & 2.56 (0.002) 
&  0.87  & 1.0   &  0.93 (0.002) & 2.09 (0.003)  
\\

KF(0.05) & \# & 0.45  & \# & 368.1 (13.21)  
& \# & 0.04  & 0.28 (0.041) & 453.9 (13.18) 
\\

KF(0.1)& 0.93  & 1.0  &0.96 (0.002) & 362.5 (13.69) 
& 0.92  & 1.0  & 0.95 (0.002) & 460.4 (13.97)
\\ 

KF(0.2)&  0.82  & 1.0 & 0.89 (0.003) & 351.7 (13.01)
&  0.80  &1.0  & 0.88 (0.003) & 455.9 (13.89) 
\\
  \hline
    &\multicolumn{8}{c}{$p=2000$, $k=10$} \\
   \hline

ET-Lasso &0.97 & 1.0  & \textbf{0.98 (0.001)} & 0.47 (0.002)
& 0.94  & 1.0  & \textbf{0.97 (0.001)} & 0.47 (0.002)
\\ 
BIC & 0.65 & 1.0 & 0.78 (0.003) & 0.17 (0.001) 
& 0.63  & 1.0  & 0.76 (0.003) & 0.16 (0.001)
\\ 
CV & 0.17  & 1.0 & 0.29 (0.003) &1.75 ( 0.007) 
& 0.17  & 1.0  & 0.29 (0.003) & 1.73 (0.009)
\\ 

ESCV & 0.95  & 1.0 & 0.97 (0.001) &3.35 (0.008) 
& 0.94  & 1.0   & 0.97 (0.001) & 3.35 (0.003)
\\

KF(0.05) & \#  & 0.002  & \# & 1252.8 (42.86) 
& \# & 0.0  & \# & 1694.6 (48.66) 
\\ 
KF(0.1)& 0.92 & 1.0  & 0.96 (0.06) & 1221.9 (41.78)
& 0.92  & 1.0  & 0.95 (0.06) & 1660.8 (47.59) 
\\ 
KF(0.2)& 0.82  & 1.0  & 0.89 (0.10) & 1200.6 (41.25)  
& 0.82  & 1.0   & 0.89 (0.10) & 1612.4 (45.91) 
\\ 

\hline
    &\multicolumn{8}{c}{$p=2000$, $k=15$} \\
 \hline
ET-Lasso & 0.98  & 1.0  & \textbf{0.99 ( 0.0006)} & 0.46 (0.002)  
&  0.95  & 1.0   & \textbf{0.97 (0.001)} & 0.46 (0.003)  
\\ 

BIC & 0.61  & 1.0  & 0.75 (0.003) & 0.16 (0.001) 
& 0.58  & 1.0   & 0.73 (0.003) & 0.16 (0.001) 
\\ 

CV & 0.17  & 1.0 & 0.29 (0.002) & 1.71 (0.008) 
&  0.17  & 1.0   & 0.29 (0.002) & 1.72 (0.009)
\\ 

ESCV & 0.88  & 1.0 & 0.94 (0.001) & 3.27 (0.006) 
&  0.86  & 1.0   & 0.92 (0.001) & 3.58 ( 0.009)
\\ 

KF(0.05) &  \#  & 0.03   &  \#  & 1251.6 (42.60)  
&  \# & 0.03  & \# & 1689.2 (48.11) 
\\ 

KF(0.1)& 0.92  & 1.0  & 0.96 (0.06)  & 1240.5 (41.71) 
& 0.93  & 1.0   & 0.96 (0.06) & 1658.4 (47.12) 
\\ 

KF(0.2)& 0.82 & 1.0 & 0.89 (0.002) & 1192.2 (40.13) 
& 0.82  & 1.0   & 0.89 (0.002) & 1610.8 (45.61) \\
\hline
\end{tabular} 
}
\end{table*}

\begin{table}[htbp]
\centering
   \caption{Simulation results of ET-Lasso, Lasso+BIC, Lasso+CV, Lasso+ESCV and Knockoff with different FDR thresholds in CS covariance structure setting. Numbers in parentheses denote the corresponding standard deviations over 1000 replicates. ($\#$ indicates an invalid MSE when zero feature is selected in some replications).}
    \label{tab:cs}
\scalebox{0.9}{
\begin{tabular}{lcccc}

         \hline
         & ${\cal P}$ & ${\cal R}$ & ${\cal F}_1$& Time  \\ 
   \hline

     &\multicolumn{4}{c}{$p=1000$, $k=10$} \\

 \hline
ET-Lasso 
& 0.89  & 1.0 & \textbf{0.93 (0.003) }& 0.26 (0.001) \\

BIC
& 0.57  & 1.0  & 0.71 ( 0.004)  & 0.09 (0.0003)\\ 

CV 
& 0.20  & 1.0  & 0.32 (0.003) & 0.94 (0.004)\\ 

ESCV 
& 0.87 & 1.0  & 0.92 (0.002) & 2.09 (0.003)\\

KF(0.05) 
& \#  & 0.00 & \# & 53.22 (0.208) \\ 

KF(0.1)
& \#  & 0.94  &  \#& 51.01 (0.211) \\ 

KF(0.2)
& 0.83  & 0.99  & 0.89 (0.003) & 50.6 (0.186)\\
   \hline

    &\multicolumn{4}{c}{$p=1000$, $k=15$} \\
            \hline

ET-Lasso
& 0.92  & 1.0& \textbf{0.95 (0.002)} & 0.26 (0.001)\\ 

BIC 
& 0.55  & 1.0  & 0.70 (0.003) & 0.09 (0.0003)  \\ 

CV
& 0.20  & 1.0  & 0.34 (0.002) & 0.93 (0.004)\\ 

ESCV
& 0.82  & 1.0  & 0.90 (0.002) & 2.07 (0.002)\\

KF(0.05) 
& \# & 0.02  & \#& 53.20(0.212) \\

KF(0.1)
& \# & 0.98  & \# & 51.12 (0.215)  \\ 

KF(0.2)
&  0.82  & 0.99  & 0.89 (0.08) & 50.67 (0.198)  \\
  \hline
    &\multicolumn{4}{c}{$p=2000$, $k=10$} \\
   \hline

ET-Lasso 
& 0.86  & 1.0  & \textbf{0.91 (0.004)} & 0.46 (0.002)\\ 
BIC 
& 0.53  & 1.0  & 0.68 (0.004) & 0.16 (0.001)\\ 
CV 
& 0.17  & 1.0  & 0.28 (0.003) & 1.63 (0.006)\\ 

ESCV 
& 0.87 & 1.0  & 0.93 ( 0.002) & 3.60 (0.016)\\

KF(0.05)
&  \#  & 0.03  & \#  & 119.1 (0.473)\\ 
KF(0.1)
& \#  & 0.79  & \#  & 115.6 (0.464)  \\ 
KF(0.2)
& \#  & 0.97  & \#  & 116.1 (0.474)\\ 

\hline
    &\multicolumn{4}{c}{$p=2000$, $k=15$} \\
 \hline
ET-Lasso 
& 0.90  & 1.0   & \textbf{0.94 ( 0.003)} & 0.45 (0.002)\\ 

BIC 
& 0.51  & 1.0  &  0.67 (0.003) & 0.16 (0.001)\\ 

CV 
& 0.17   & 1.0  & 0.29 (0.002) & 1.63 (0.007)\\ 

ESCV 
& 0.76   & 1.0  & 0.86 (0.003) & 3.67 (0.017)\\ 

KF(0.05)
& \# & 0.02  & \#  & 119.8 (0.522)\\ 

KF(0.1)
& \# & 0.93 & \# & 116.2 (0.494) \\ 

KF(0.2)
& \#  & 0.96  & \#  & 115.9 (0.495)\\
   \hline
    \end{tabular}
    }
\end{table}

\subsection{FIFA 2019 Data}
In this experiment, the ET-Lasso method is applied to a FIFA 2019 dataset, which is a random sample from a Kaggle data \cite{fifa}. The dataset contains 84 attributes of 2019 FIFA complete players. We have 1500 players in this example. The response variable is the wage of each player, and the rest 83 attributes are feature candidates that may affect a player's wage. We standardize the response and the features. The training data consists of 300 players and the rest 1200 players are used for testing. The mean squared error (MSE), the number of selected features (DF) of different methods based on $100$ replications are reported in Table \ref{tab:fifa}. 

\begin{table}
	\begin{center}
		\caption{\label{tab:fifa} Comparison of ET-Lasso, Lasso+CV, Lasso+ESCV, Lasso+BIC and Knockoff (KF) on FIFA 2019 Data.}
		
		\scalebox{1}{%
			\begin{tabular}{cccc}
				\hline
				
				&MSE& DF & Time\\ 
				\hline
 \textbf{ET-Lasso}&0.2228 (0.0027) & 4.92 (0.224) & 0.0958 (0.004) \\
				CV&0.2277 (0.0034) & 11.6 (0.880) & 0.2841 (0.007)\\
				BIC&0.2320 (0.0036) & 1.86 (0.086) &0.0211 (0.007)\\
				ESCV&0.2408 (0.0080) & 11.6 (0.880) &0.8195 (0.007) \\
				KF(0.1)& \# & 0.2 (0.141)&  0.9554 (0.014)\\
				KF(0.3) & \# & 2.94 (0.331)& 0.9698 (0.014)\\ 
				KF(0.5) & \# & 4.82 (0.248) & 0.9692 (0.017) \\
				\hline							
			\end{tabular}
		}
	\end{center}
\end{table}
We compare ET-Lasso with Lasso+CV, Lasso+ESCV, Lasso+BIC and Knockoff (KF). Since Knockoff cannot estimate $\beta$ directly, we implement a two stage method for Knockoff, where at the first stage we apply Knockoff for feature selection, and at the second stage, we apply linear regression model with selected features and make predictions on test data. From Table \ref{tab:fifa}, we can see that ET-Lasso has the best MSE with the smallest standard deviation among all methods. Specifically, ET-Lasso improves the prediction MSE of CV and ESCV by around 2\% and 7\%, while its model size is on average less than half of the latter two. ET-Lasso takes less than 0.1s to run, which is significantly more efficient than CV and ESCV. ET-Lasso improves the MSE of BIC by more than 4\%. BIC on average selects less than two features into the model, which might indicate an underfitting. For this real data, Knockoff-based methods are too aggressive to select any feature during some replications, which makes their MSE uncomputable, even when the FDR threshold is set as large as $0.5$. 

\subsection{Stock Price Prediction}
\begin{figure*}
	\centering
	\includegraphics[width=13cm, height=6.3cm]{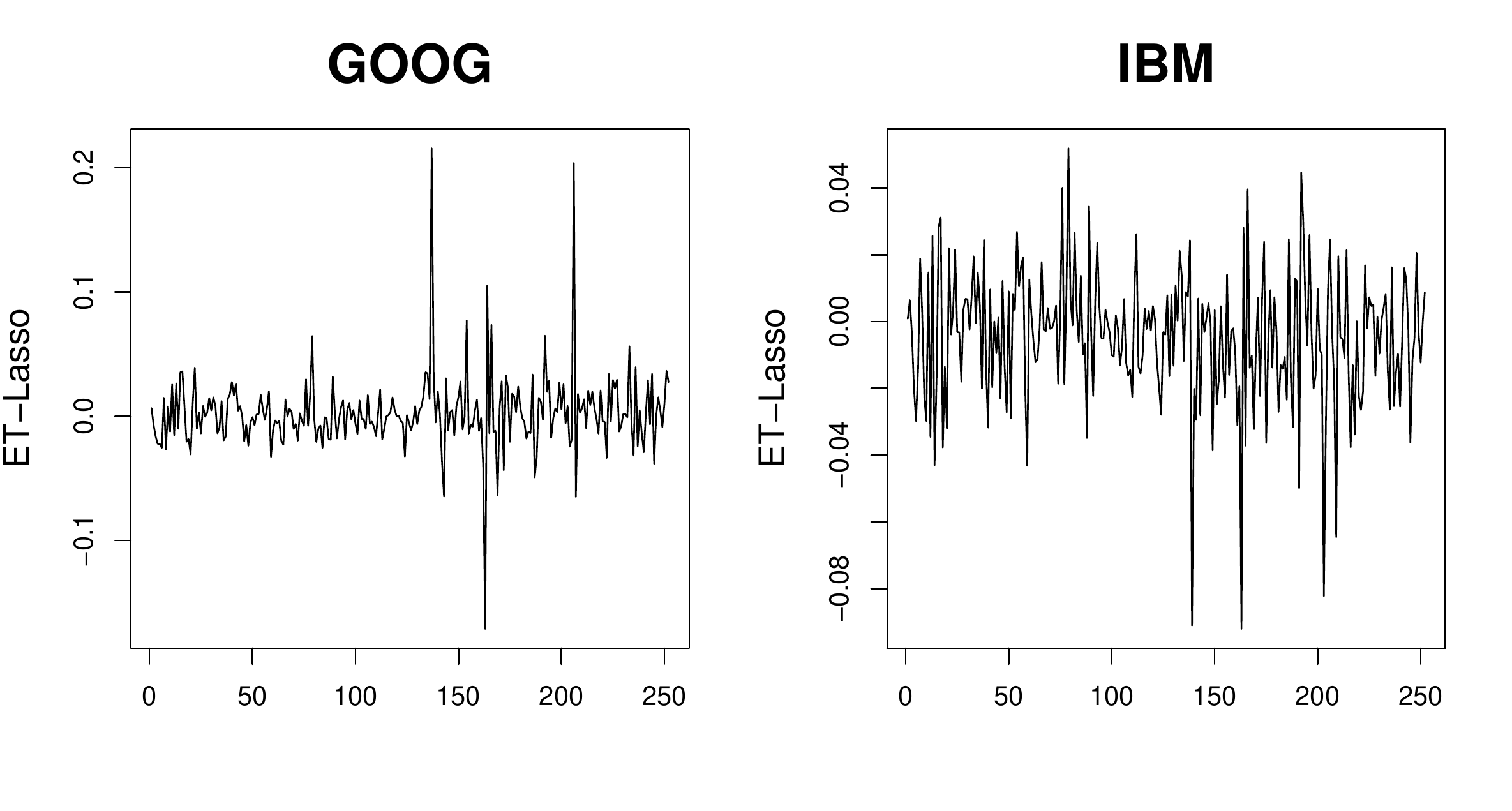}
	\includegraphics[width=13cm, height=6.3cm]{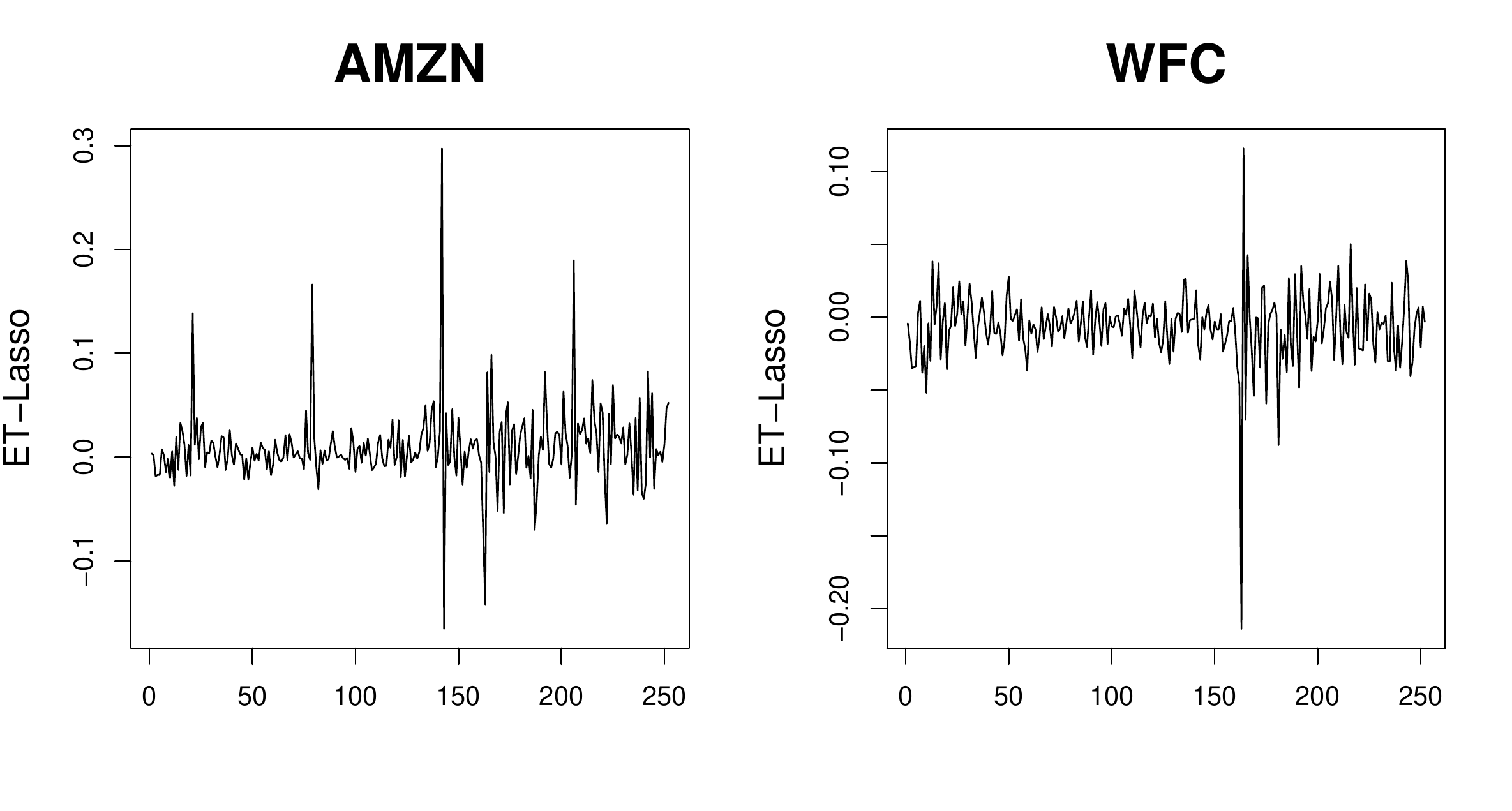}
	\caption{Difference of the true stock price and the predicted stock price. }
	\label{fig:stock}
\end{figure*}

In this example, we apply the ET-Lasso method for stock price prediction. We select four stocks from four big companies, which are GOOG, IBM, AMZN and WFC. We plan to use the stock open price from date 01/04/2010 to 12/30/2013 to train the model, and then predict the open price in the trading year 2014. \textbf{All the stock prices are normalized}. Considering that the current open price of a stock might be affected by the open price of the last 252 days (number of trading days in a year), we apply the following regression model,
\beq
 y_t = a + \sum\limits_{i=1}^{252}\beta_{i}y_{t-i}, 
\eeq
and the coefficients estimator is obtained as
\beq 
\hat{\bbeta}(\lambda)=\argminB_{\bbeta}\frac{1}{2n}\|\by-\sum\limits_{i=1}^{252}\beta_{i}y_{t-i}\|^2_2+\lambda\sum\limits_{j=1}^{252}|\beta_j|.
\eeq
Figure \ref{fig:stock} depicts the difference of predicted price using ET-Lasso and the true price, i.e., $y_t-\hat{y}_t$ for the four stocks. 
It can be seen that ET-Lasso method predicts the trend of the stock price change very well. The mean squared error (MSE) and the number of selected features (DF) are reported in Table \ref{tab:stock}. We can observe that the ET-Lasso method outperforms Lasso+BIC and Lasso+CV in terms of both prediction error and model complexity. For instance, when we predict the stock price of WFC, the MSE of ET-Lasso method is $6.45\times 10^{-4}$, which is only about $1/10$ of that of Lasso+CV ($5.9\times 10^{-3}$)  and about $1/100$ of that of Lasso+BIC ($6.78\times 10^{-2}$). Knockoff methods with a controlled FDR smaller than 0.5 are over-aggressive in feature selection, leading to an empty recovery set in most circumstances. KF(0.5) works well on IBM, AMZN and WFC, with resulting MSE comparable to that of ET-Lasso; however it selects zero feature on GOOG stock. In terms of the computing efficiency, ET-Lasso is much faster than Knockoff and cross-validation method and a bit slower than BIC.

\begin{table}[htp]
	\begin{center}
		\caption{\label{tab:stock} Comparison of ET-Lasso, Lasso+CV, Lasso+ESCV, Lasso+BIC and Knockoff (KF) on Stock Price Prediction.}
		\scalebox{0.9}{%
			\begin{tabular}{ccccccc}
				\hline
				&\multicolumn{3}{c}{GOOG}&\multicolumn{3}{c}{IBM}\\  \hline
		       & MSE &DF&Time& MSE &DF&Time\\ 
		       \hline
				ET-Lasso&$9.40\times 10^{-4}$& 9&0.25 &$4.29\times 10^{-4}$&3& 0.12 \\
			    CV&$9.68\times 10^{-4}$& 9&0.57&$4.59\times 10^{-4}$& 3&0.29\\
				BIC&$2.43\times 10^{-2}$&2&0.06  &$5.81\times 10^{-4}$& 2&0.02\\
				ESCV&$9.47\times 10^{-4}$& 9&1.2 &$4.37\times 10^{-4}$&3& 0.62 \\
				
				KF(0.1)& \# & 0&8.26 &\#&0& 8.48 \\
				KF(0.3)& \# & 0&8.24 &\#&0& 7.71 \\
				KF(0.5)& \# & 0&7.57 &$4.38\times 10^{-4}$&4 & 8.74 \\ 
				\hline	
				&\multicolumn{3}{c}{AMZN}&\multicolumn{3}{c}{WFC}\\
				\hline
				&MSE &DF&Time & MSE &DF &Time \\
				ET-Lasso&$1.54\times 10^{-3}$&8 &0.15 & $6.45\times 10^{-4}$&11 &0.16 \\
				CV&$1.63\times 10^{-3}$&9& 0.43 & $5.90\times10^{-3}$&11 &0.64 \\
				BIC&$7.55\times 10^{-3}$&2 & 0.05 & $6.78\times 10^{-2}$ & 3& 0.05 \\
				ESCV&$1.57\times 10^{-3}$&8 &0.90 & $8.67\times 10^{-4}$&11 &0.16 \\
				KF(0.1)&\#&0&7.85  & \# &0& 9.19 \\
				KF(0.3)&\#&0& 8.05 & \# &0& 8.06\\
				KF(0.5)&$1.55\times 10^{-3}$&6 & 7.97& $6.20\times 10^{-4}$ &6& 7.92  \\
				\hline

			\end{tabular}
		}
	\end{center}
\end{table}

\subsection{Chinese Supermarket Data}
In this experiment, the ET-Lasso method is applied to a Chinese supermarket dataset in \cite{wang2009forward}, which records the number of customers and the sale volumes of $6398$ products from year 2004 to 2005. The response is the number of customers and the features include the sale volumes of 6398 products. It is believed that only a small proportion of products have significant effects on the number of customers. The response and the features are both standardized. The training data includes the 60\% days and the rest is used as testing data. The mean squared error (MSE), the number of selected features (DF) of the ET-Lasso method, cross-validation (CV), BIC and Knockoff (KF) are reported in Table \ref{tab:market}. 
\begin{table}[htp]
	\begin{center}
		\caption{\label{tab:market} Comparison of ET-Lasso, Lasso+CV, Lasso+ESCV, Lasso+BIC and Knockoff (KF) on Chinese supermarket data.}
		
		\scalebox{0.9}{%
			\begin{tabular}{cccc}
				\hline
				
				&MSE& DF & Time\\ 
				\hline
				\textbf{ET-Lasso}&0.1046 &68 &1.40\\
				CV&0.1410&111& 5.80\\
				BIC&0.3268&100&0.517\\
				ESCV&0.1172&72& 12.60\\
				KF(0.1)& \# & 0 & 1449.355  \\
				KF(0.3)& 0.1465 & 11 & 1358.877 \\
				KF(0.5)& \# & 0 & 1379.757 \\
				\hline							
			\end{tabular}
		}
	\end{center}
\end{table}

We can see that ET-Lasso performs best with respect to the model prediction accuracy. ET-Lasso method returns the smallest prediction MSE (0.1046) and a simpler model (contains 68 features) than CV and BIC. Cross-validation and BIC for Lasso lead to larger MSE and model size. For the Knockoff method, when FDR is controlled as small as $0.1$ or as large as $0.5$, it fails to select any feature. Knockoff with $0.3$ FDR only selects $11$ features, but the prediction MSE is relatively large. Knockoff-based methods take more than $1000$ seconds to run, which are significantly slower than ET-Lasso (1.4s), Lasso+CV (5.8s), Lasso+ESCV (12.5s) and Lasso+BIC (0.517s).

\begin{figure}
	\centering
	\includegraphics[width=7cm, height=6cm]{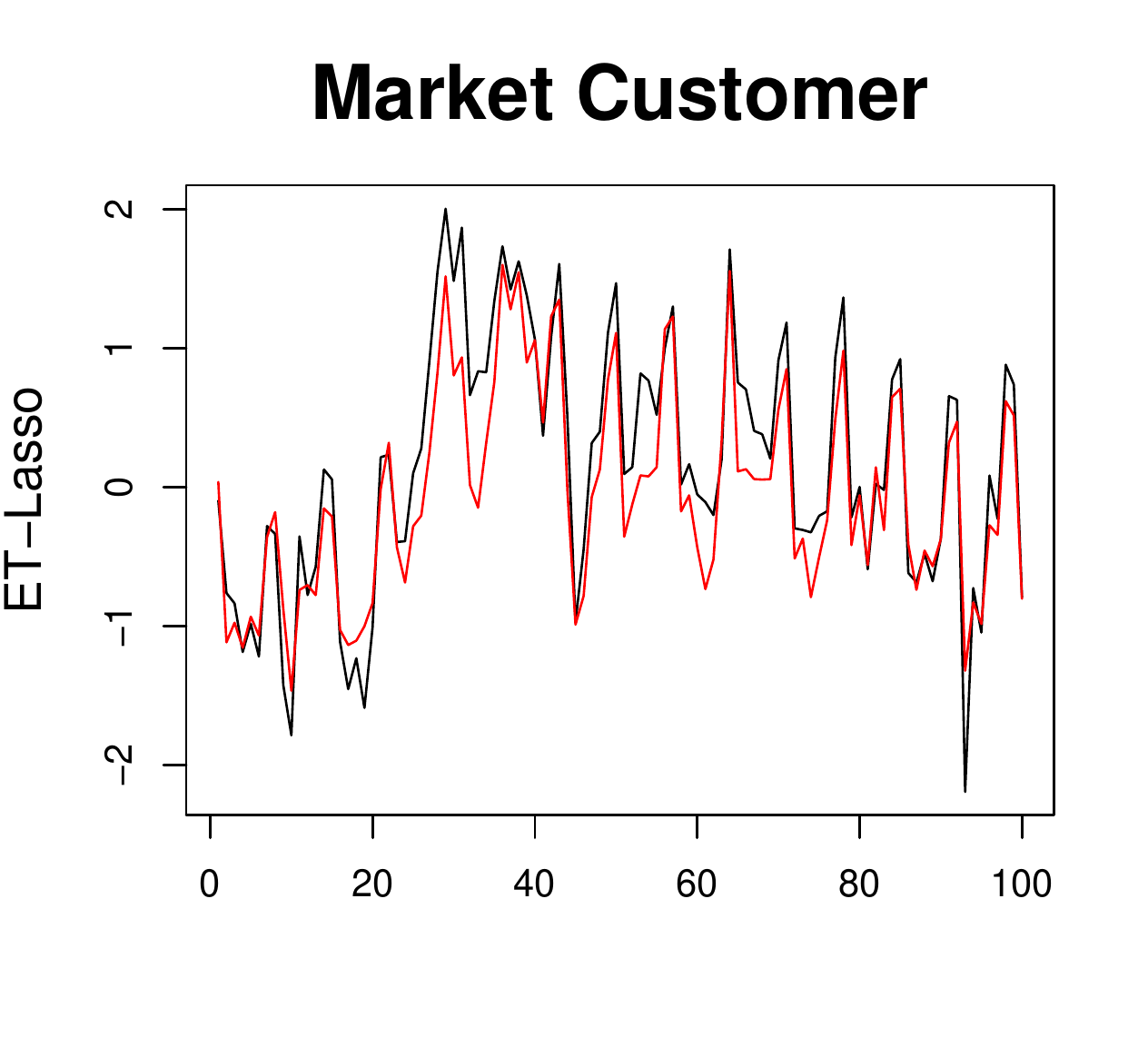}
	\caption{Performance of the ET-Lasso method on Chinese supermarket Data. The black line is the true value and the red line is the predicted value. }
\end{figure}

\subsection{Extension to Classification}

In this part, we test the performance of ET-Lasso for a binary classification problem. Although the theoretical foundation of ET-Lasso is based on the linear regression model setting, we want to show that ET-Lasso also performs well in more general cases. Here we provide one example of its extension to logistic regression. 

The dataset comes from the UCI machine learning repository, \emph{Smartphone Dataset for Human Activity Recognition (HAR) in Ambient Assisted Living (AAL)} \footnote{\url{https://archive.ics.uci.edu/ml/datasets/Human+Activity+Recognition+Using+Smartphones}}.
The dataset was collected from an experiment in the area of Ambient Assisted Living. In the experiment, 30 participants from 22 to 79 years old were asked to wear smartphones around their waists. The smartphones have built-in accelerometer and gyroscope. For each participant, six activities (standing, sitting, laying, walking, walking upstairs, walking downstairs) were performed for one minute, and the 3-axial raw signals from accelerometer and gyroscope were collected. Based on these signals, $561$-feature vectors were extracted and derived as potential predictors for activity recognition. We select records corresponding to two activities, walking and standing, and aim to identify relevant features that can differentiate between them. We random sample $200$ and $100$ observations from the given training set as two training datasets. The testing data contains $497$ records. Since the data is relatively balanced, we report the average number of misclassification and the number of selected features (DF) over $100$ replications in Table \ref{tab:har}. It can be seen that ET-Lasso has the best classification accuracy among all methods. Knockoff-based methods are too aggressive for this task. The performances of BIC and CV are more sensitive to the sample size than ET-Lasso, and ESCV might exclude some of the active features.

\begin{table}[htp]
	\begin{center}
		\caption{\label{tab:har} Comparison of ET-Lasso, Lasso+CV, Lasso+ESCV, Lasso+BIC and Knockoff (KF) on Human Activity Recognition Data.}
		
		\scalebox{0.95}{%
					
		\begin{tabular}{ccccc}
				\hline
				
	&	& Number of Misclassification  & DF \\ 
				\hline
$n=200$ & \textbf{ET-Lasso}   & 5.6 (0.241)& 10.5 (0.395)  \\
&				CV   & 5.7 (0.226)         & 12.5 (0.417) \\
&				BIC  & 6 (0.291)           & 6 (0.226)     \\
&				ESCV & 8.7 (0.294)         & 4.5 (0.417)  \\
&				KF(0.1) & \# & 0\\
&				KF(0.3) & \# & 0\\ 
&				KF(0.5) & \# & 0\\
				\hline							

				
				\hline
	$n=100$	 & \textbf{ET-Lasso}   & 5.7 (0.365)&9.2 (0.524)  \\
&				CV   & 6.5 (0.435) & 12.9 (0.508) \\
&				BIC  & 6.6 (0.403) & 4.5 (0.117)     \\
&				ESCV & 8.2 (0.459) & 4.1 (0.202)  \\
&				KF(0.1) & \# & 0\\
&				KF(0.3) & \# & 0\\ 
&				KF(0.5) & \# & 0\\
				\hline							
			\end{tabular}

		}
	\end{center}
\end{table}

\section{CONCLUSION}
\label{sec:conclusion}
In this paper, we propose ET-Lasso that is able to select the ideal tuning parameter by involving pseudo-features. The novelties of ET-Lasso are two-fold. First, ET-Lasso is statistically efficient and powerful in the sense that it can select all active features with the smallest model which contains least irrelevant features (i.e., highest precision) compared to other feature selection methods. Second, ET-Lasso is computationally scalable, which is essential for high-dimensional data analysis. The ET-Lasso is efficient for tuning parameter selection of regularization methods and requires no calculations of the prediction error and posterior model probability. Moreover, ET-Lasso is stopped once the cutoff is found, so there is no need to traverse all potential tuning parameters as cross-validation and BIC. On the other hand, Knockoff turns out to be very computational intensive for high dimensional data. Numerical studies have illustrated the superior performance of ET-Lasso over the existing methods under different situations. 


\section{ACKNOWLEDGMENTS}
We want to sincerely thank Dr. Runze Li and Dr. Bharath Sriperumbudur 
for their insightful and constructive comments on this work.

\bibliographystyle{acm}
\bibliography{sample-base}

\clearpage

\newpage
\section*{\textbf{Supplement}}
\subsection*{Proofs of Theorem 1}
To prove Theorem 1, we need to find a cut off which separate ${Z_j, j \in s^*}$ and ${Z_j, j \in s^*_c}$. In order to do this, we first introduce Lemma \ref{lem1}. 
\begin{lemma}\label{lem1}
	Under conditions C1-C3, if
	\beq  \label{5}
	\lambda = O(n^{\delta_3})
	\eeq
	for some $\delta_3 > (\delta_1-1)/2$, then the problem (2.2) has a minimizer where $\hat{\bbeta}_{s^*_c}=0$ with probability going to $1$ as $n \rightarrow \infty$.
	In addition, if $\min_{j \in s^*}|\beta_j|^2> (\lambda[\frac{2\sigma}{\sqrt{C_{\min}}}+\sqrt{k}/ C_{\min}])^2$, then $sign(\hat{\bbeta}_{s^*})=sign(\bbeta_{s^*}^*)$, where $sign(\bx)$ is the sign function of $\bx$ and $\bbeta_{s^*}^*$ is the true coefficients of important predictors.  	
\end{lemma}
To prove Lemma 1, we need the following Lemma 2. We first define $P_{\lambda}(|\bbeta_{s^*} |) = \lambda\sum\limits_{j=1}^{p}|\beta_{j} |$ .
\begin{lemma}\label{lem2}
	Under conditions C1-C3, suppose $\hat{\bbeta}= (\hat{\bbeta}_{s^*}, \hat{\bbeta}_{s^*_c})$, where $\hat{\bbeta}_{s^*}$ is a local minimizer of the restricted PLS problem
	\beq \label{2}
	\hat{\bbeta}_{s^*}=\argminB_{\bbeta_{s^*} \in R^k}\{\frac{1}{2n}\|\by-\bX_{s^*} \bbeta_{s^*} \|^2_2+P_{\lambda}(|\bbeta_{s^*} |) \}, 
	\eeq
	and $\hat{\bbeta}_{s^*_c}=0$. If
	\beq \label{3}
	||n^{-1}\bX_{s^*_c}^T(\by-\bX_{s^*}\hat{\bbeta}_{s^*})||_{\infty}<\lambda,
	\eeq
	then $\hat{\bbeta}$ is a local minimizer of (2.2),
	and if 
	\beq\label{6.4}
	sign(\hat{\bbeta}_{s^*})=sign(\bbeta_{s^*}^*),
	\eeq 
	then $\hat{\bbeta}$ is a local minimizer of (2.2) with correct signed support. 
\end{lemma}

\noindent\textbf{Proof of Lemma \ref{lem2}.}
Under conditions C1-C3 and (\ref{3}) is satisfied, it is trival to conclude that $\hat{\bbeta}$ is a local minimizer of the Penalized Likelihood problem (2.2) based on the Theorem 1 proposed in \cite{fan2011nonconcave}. If (\ref{6.4}) is also satisfied, then $\hat{\bbeta}$ is a local minimizer of regularization problem with correct signed support. 

\noindent\textbf{Proof of Lemma \ref{lem1}.} 
Construct $\hat{\bbeta}$ as in Lemma \ref{lem2}. First, we prove (\ref{3}). 
where $\partial P_{\lambda}(x)$ is the sub-gradient of $P_{\lambda}(x)$:
\beq
\partial P_{\lambda}(|x|)
\left\{
\begin{array}{cll}
	\in & (-\lambda, \lambda)& x=0 \\
	= & P_{\lambda}^{'}(|x|)sign(x)& otherwise.
\end{array}
\right.
\eeq
\beq
\begin{aligned}
n^{-1}\bX_{s^*_c}^T(\by-\bX_{s^*}\hat{\bbeta}_{s^*}) &=n^{-1}\bX_{s^*_c}^T[ \bX_{s^*}(\frac{1}{n}\bX_{s^*}^T\bX_{s^*})^{-1}\partial P_{\lambda}(|\hat{\bbeta}_{s^*}|)\\
&+(I-\bX_{s^*}(\bX_{s^*}^T\bX_{s^*})^{-1}\bX_{s^*}^T) \beps],
\end{aligned}
\eeq
Then it follows that 
\beq
\begin{aligned}
\|n^{-1}\bX_{s^*_c}^T(\by-\bX_{s^*}\hat{\bbeta}_{s^*})\|_{\infty} &\leq \|\bX_{s^*_c}^T \bX_{s^*}(\bX_{s^*}^T\bX_{s^*})^{-1}\partial P_{\lambda}(|\hat{\bbeta}_{s^*}|)\|_{\infty}\\
&+\|n^{-1}\bX_{s^*_c}^T(I-\bX_{s^*}(\bX_{s^*}^T\bX_{s^*})^{-1}\bX_{s^*}^T) \beps\|_{\infty}.
\end{aligned}
\eeq

Under conditions C1 and C3, the first term \\
$ \|\bX_{s^*_c}^T \bX_{s^*}(\bX_{s^*}^T\bX_{s^*})^{-1}\partial P_{\lambda}(|\hat{\bbeta}_{s^*}|)\|_{\infty}$ is controlled by $(1-\gamma)\lambda$. 

$\forall j \in {s^*_c}$, 
\[
\|n^{-1}\bX_{j}^T(I-\bX_{s^*}(\bX_{s^*}^T\bX_{s^*})^{-1}\bX_{s^*}^T) \beps\|_{\infty}\leq \|\frac{1}{n}\bX_j^T \beps\|_{\infty},  
\]	
based on the fact that the projection matrix $(I-\bX_{s^*}(\bX_{s^*}^T\bX_{s^*})^{-1}\bX_{s^*}^T)$ has spectral norm one. In addition, if $n^{-1/2}\max_j \|\bX_j\|_2 \leq 1$, the sub-Gaussian tail bound satisfies
\[
P(\|\frac{1}{n}\bX_j^T \beps\|_{\infty}\geq \gamma \lambda) \leq 2(p-k)\exp(-\frac{n\gamma^2\lambda^2}{2\sigma^2}).
\] 
Thus, 
\[
P(||n^{-1}\bX_{s^*_c}^T(\by-\bX_{s^*}\hat{\bbeta}_{s^*})||_{\infty}<\lambda) \geq 1-2\exp(-\frac{n\gamma^2\lambda^2}{2\sigma^2}+\log(p-k)). 
\]
By condition C3, 
\[
P(||n^{-1}\bX_{s^*_c}^T(\by-\bX_{s^*}\hat{\bbeta}_{s^*})||_{\infty}<\lambda )>1-2\exp(-C_1 n^{1+2\delta_3}),
\]
so the probability goes to $1$ as $n\longrightarrow \infty$. 

In the second step, we want to know when the sign consistency of $\hat{\bbeta}$ can be guaranteed. The difference between $\bbeta_{s^*}^*$ and $\hat{\bbeta}_{s^*}$ is 
\[
\Delta=\bbeta_{s^*}^*-\hat{\bbeta}_{s^*}=(\frac{1}{n}\bX_{s^*}^T\bX_{s^*})^{-1}[\partial P_{\lambda}(|\hat{\bbeta}_{s^*}|)-\frac{1}{n}\bX_{s^*}^T\beps].
\]
So the $\max_i \Delta_i$ is bounded by 
\[
\|(\frac{1}{n}\bX_{s^*}^T\bX_{s^*})^{-1}\partial P_{\lambda}(|\hat{\bbeta}_{s^*}|) \|_{\infty}+\|(\frac{1}{n}\bX_{s^*}^T\bX_{s^*})^{-1}\frac{1}{n}\bX_{s^*}^T\beps\|_{\infty}.
\]

The first term is a deterministic quantity, therefore, we only need to bound the second term. For $i=1, \cdots, k$, define $d_i$ by
\[
d_i=\bee_i^T (\frac{1}{n}\bX_{s^*}^T\bX_{s^*})^{-1}\frac{1}{n}\bX_{s^*}^T\beps. 
\]
where $\bee_i$ is a p-vector with $ith$ element equal to $1$ and other elements equal to $0$. By condition C2, we have 
\[
\frac{\sigma^2}{n}\|(\frac{1}{n}\bX_{s^*}^T\bX_{s^*})^{-1}\|_2\leq \frac{\sigma^2}{n C_{\min}}.
\]

Based on the property of sub-Gaussian random variable, it can be derived that
\[
P(\|(\frac{1}{n}\bX_{s^*}^T\bX_{s^*})^{-1}\frac{1}{n}\bX_{s^*}^T\beps\|_{\infty}>2\sigma \lambda/\sqrt{C_{\min}})\leq 2\exp(-\frac{2n \lambda^2}{\sigma^2}+\log k).
\] 

In addition, since $\bX_{s^*}^T \bX_{s^*}$ is a $k \times k$ matrix, we have
\[
\lambda\|(\frac{1}{n}\bX_{s^*}^T\bX_{s^*})^{-1}\|_{\infty}\leq \lambda\sqrt{k} \|(\frac{1}{n}\bX_{s^*}^T\bX_{s^*})^{-1}\|_{2}\leq \lambda\sqrt{k}/ C_{\min}.
\]
Therefore,   
\[
\|\hat{\bbeta}_{s^*}-\bbeta_{s^*}^*\|_{\infty}\leq \lambda[\frac{2\sigma}{\sqrt{C_{\min}}}+\sqrt{k}/ C_{\min}]
\]
with a probability larger than $1-2\exp(-c_2 n^{1+2\delta_{3}})$. Besides, we know that $\min_{j\in S}|\beta_j|^2>(\lambda[\frac{2\sigma}{\sqrt{C_{\min}}}+\sqrt{k}/ C_{\min}])^2$, so $sign(\hat{\bbeta})=sign(\bbeta^*)$ with probability going to $1$ as $n\rightarrow\infty$ under conditions C1-C3. Thus Lemma \ref{lem1} is proved.

\noindent\textbf{Proof of Theorem 1.} 
From Lemma \ref{lem1}, it is obvious that when $\min_{j\in {s^*}}|\beta_j^*|> O(n^{\delta_1+\delta_2-1})$, $\min_{j\in s^*} Z_j > O(n^{(\delta_1-1)/2})$, whereas $\max_{j\in s^*_c} Z_j \leq O(n^{(\delta_1-1)/2})$ with probability going to $1$ as $n \rightarrow \infty$. Therefore, Theorem 1 is proved.



\end{document}